\documentclass[10pt,twocolumn,letterpaper]{article}

\usepackage{cvpr}
\usepackage{times}
\usepackage{epsfig}
\usepackage{graphicx}
\usepackage{amsmath}
\usepackage{amssymb}

\usepackage[breaklinks=true,bookmarks=false]{hyperref}

\cvprfinalcopy % *** Uncomment this line for the final submission

\def\cvprPaperID{****} % *** Enter the CVPR Paper ID here

\begin{document}

%%%%%%%%% TITLE
\title{CAAD 2018: Powerful None-Access Black-Box Attack Based on \\Adversarial Transformation Network}
\author{Xiaoyi Dong\\
USTC\\
{\tt\small dlight@mail.ustc.edu.cn}
\and
Weiming Zhang\\
USTC\\
{\tt\small zhangwm@ustc.edu.cn}
\and
Nenghai Yu\\
USTC\\
{\tt\small 	ynh@ustc.edu.cn}
}


\maketitle
%\thispagestyle{empty}

%%%%%%%%% BODY TEXT

\begin{figure*} 
\centering
\includegraphics[width=6in,height=1.5in]{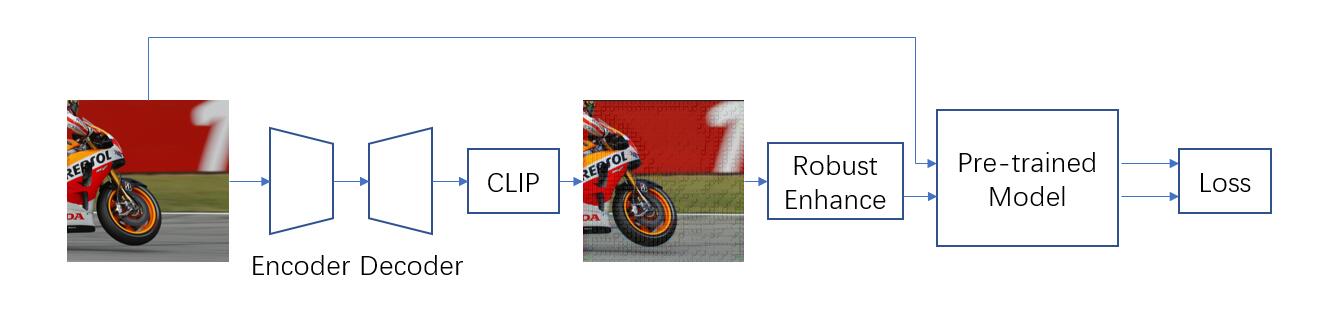}
\caption{Architecture of our method}
\label{fig:arc}
\end{figure*}
\section{Method}
In this paper, we propose an improvement of Adversarial Transformation Networks(ATN) \cite{DBLP:journals/corr/BalujaF17}to generate adversarial examples, which can fool white-box models and black-box models with a state of the art performance and won the \textbf{SECOND} place in the non-target task in CAAD 2018. In this section, we first introduce the whole architecture about our method, then we present our improvement on loss functions to generate adversarial examples satisfying the $L_{\infty}$ norm restriction in the non-targeted attack problem. Then we illustrate how to use a robust-enhance module to make our adversarial examples more robust and have better transfer-ability. At last we will show our method on how to attack an ensemble of models. 
\subsection{Model Architecture}
Our work is based on ATN and propose a new training framework and two powerful loss functions for improving the transfer-ablity and training speed. Figer~\ref{fig:arc} shows our framework \par
Our framework is composed by a Generate module and a Robust-enhance module. In the Generate module, there are the Encoder and Decoder just like the ATN. But before feeding the clean image and adversarial example into the pre-trained model, we add a Robust-enhance module to imitate the image pre-process used in some defence methods.\par
The generating part could be defined as a neural network:\par
\begin{equation}
    g_{k,\theta}(x): x \in \chi \to x'
\end{equation}
where $\theta$ is the parameter of $g(x)$, $k(x)$ is the target model which outputs a probability distribution across class labels and $k_f(x)$ is the feature map before the target model's last average pooling layer. $x$ is the input image and $\chi$ is the distribution domain of the input image, $x'$ is the adversarial example generate by the $g(x)$ and 
\begin{equation}
    argmax \; k(x) \ne argmax \; k(x')
\end{equation}

\subsection{Loss Function}
%As the competition only concern about the $L_{\infty}$ perturbation, we abandon the space-domain loss and use the following loss functions to improve the performance.\\
With the $L_{\infty}$ norm restriction, we abandon the space-domain loss and use the following loss functions to improve performance.\\
\textbf{Feature Based Loss function}: inspired by the SRGAN \cite{DBLP:journals/corr/LedigTHCATTWS16} and Guided denoise \cite{DBLP:journals/corr/abs-1712-02976} , we use the target model's feature map before the last average pooling layer as the input images' feature and try to maximum the  $L_{1}$ distance between the feature of real image and the adversarial example.\par
\begin{equation}
  l_{F}(x,x') = 1-L_1(k_f(x),k_f(x'))
\end{equation}\par
We use this feature level loss function in the DLight team's attack model and got the third prize.\\
%As the feature based is only suitable for the non-target attack, we use a easy but powerful loss based on the output of the target model for target attack and non-target attack.\par
\textbf{Prediction Based Loss function}: we also concerned about the loss based on the target model's output and found a simple but powerful loss function. \par
\begin{equation}
  \begin{split}l_P(x,x') = 
  \begin{cases}
  P'_{fir} - P'_{sec}, & \text{if } argmax \; f(x)= fir \\
  P'_{sec} - P'_{fir}, & \text{if } argmax \; f(x) \ne  fir
  \end{cases}\end{split}\end{equation}
where for target model $f$ with $N$ class, it's output prediction $f(x')=[P'_1,P'_2, \ldots P'_{N}]$, we define $fir$ and $sec$ as labels of the top two probabilities in $f(x')$ and $P_{fir}$ and $P_{sec}$ are their corresponding probabilities. In the following experiments we will show that adversarial examples generated by the model trained with this simple loss have strong transfer-ability when comparing with other methods.\par
We use this loss function in the Hooin Zira's submission and won the second place in the Non-target attack competition.\par
\subsection{Robust-enhance Module}
%In recent research, most paper only concern about the transfer-ability to other black box model or the model with adversarial training, but few pay concentrates on the the image pre-process on the input image, so we used a Robust-enhance part to improve the robustness of our adversarial examples and attack these kind of defence methods. \par
As mentioned in 1.3, there are two main methods for defending adversarial examples: adversarial training and image pre-process. In order to attack the second defence methods, we insert a Robust-enhance module in the training process, after we get the output adversarial examples with clip, we add this Robust-enhance module on the adversarial examples and feed the processed images to the following target model. We find that with this Robust-enhance module, our adversarial examples are more robust to the second sort defence methods and get more powerful transfer-ability when attacking models trained with the adversarial examples(first sort defence methods). \par
Considering about the implement of back propagation, we use three basic image process in the Robust-enhance module.\\
\textbf{Random Noise}: Add random noise. \\
\textbf{Pre-trained Filter}: We randomly use the Random Noise or a small pre-trained filter network to process the adversarial examples. \\
\textbf{Training Filter}:We randomly use the Random Noise or a small pre-trained filter network to process the adversarial examples, during training the Generate module, we also train the filter. \par
%As the de-noise filter mentioned before is pretrianed and all the parameters are fixed, the Generate module may overfit the filter and have a poor performance. So we propose two method to solve this problem: first, we randomly choose pre-process method during the training process. Second, Inspired by GAN, we train the filer to improve the denoise performance during training, just like the Discriminater in GAN. We define this as \textbf{Maxmin Filter architecture} \par
\begin{table*}
\begin{center}
\begin{tabular}{|l|c|c|c|c|c|c|c|c|c|}
\hline
Attack & Inc-v3 &Inc-v4  & IncRes-v2 & PolyNet & NasNet & Res-101& Inc-v3$_{ens3}$ & Inc-v3$_{ens4}$& Mean \\
\hline\hline
FGSM    & 0.71          &0.24       & 0.23 & 0.37 &0.18 &0.34& 0.13    &0.11 & 0.22 \\
PGD     & \textbf{0.99} &0.18       & 0.12 & 0.22 & 0.09 &0.18& 0.11    &0.07 & 0.14\\
MI-FGSM & \textbf{0.99} &0.42       & 0.38 & 0.46 & 0.23 & 0.44& 0.13    &0.11 & 0.31\\

F-ATN(No Robust)    & 0.91 & 0.63  & 0.59 & 0.69 & 0.47 & 0.65&0.28  & 0.27 & 0.51\\
F-ATN   & 0.91 & 0.69  & 0.6 & 0.74 & 0.51 &0.76 & 0.30 & 0.29 & 0.55\\
P-ATN(No Robust)    & 0.98 &0.97   & 0.92 & 0.96 & 0.91 & 0.93  & 0.58 & 0.47 & 0.82\\
P-ATN   & 0.97          &\textbf{0.97} & \textbf{0.93} & \textbf{0.96} & \textbf{0.89} &\textbf{0.94} & \textbf{0.83} & \textbf{0.80} & \textbf{0.90} \\

\hline
\end{tabular}
\end{center}
\caption{Results for single model attack. Inc-v3 is the target white-box model and others are black-box models. We compare our methods P-ATN, F-ATN with FGSM, PGD and MI-FGSM, we find our method have a similar performance comparing with start of the art method MI-FGSM on white-box attack and have a much higher fooling rate when attacking black-box models. We also test the performance of Robust-enhance module, P-ATN(No Robust) and F-ATN(No Robust) show the fooling rate of model without Robust-enhance module during training, results shows that with Robust-enhance module, P-ATN's performance have a great progress when attacking black-box models, but F-ATN only have a small improvement. The last row shows the average black-box fooling rate, P-ATN have the best performance}
\label{tab:single}
\end{table*}
\subsection{Attacking ensemble of models}
In this section, we show how to attack an ensemble of models. Previous researches and competitions show that ensemble methods is a efficient method for enhancing performance and improve robustness in many area including adversarial examples. Adversarial examples generated by ensemble training are able to fool many white-box models at the same time and have better transfer-ability to attack black-box models.\par
We propose to attack multiple models by adding the loss together and we call this $ensemble \; in\;  loss$. As our loss functions are defined in both feature level and prediction level, this ensemble in loss method could easily used with any kind of loss functions. When it comes to the back propagation step, every loss will calculate gradients individually and sum at each parameter. This will guide the model to learn how to be more aggressive to all the target models. Specifically, to attack an ensemble of $N$ models, we fuse the losses as \\
\begin{equation}
    l(x)=\sum^{N}_{n=1} {w_{n}l_{n}(x)}
\end{equation}
where $l_{n}(x)$ are the loss function of the n-th model,  the loss function here could be the feature based loss function or the prediction based loss function $w_n$ is the ensemble weight with $w_n > 0$, we use it to keep balance of the gradients' magnitude from different models. \par
Meanwhile, as show in Fig.1, we find when our adversarial examples are adversarial to a model, it will be more adversarial in a few iterations. In order to forbid our generate model tend to one of the target models, we add a threshold $\gamma$ for our prediction based loss $l_P$ and get new loss $l_{ensP}(x,x')$:
\begin{equation}
  l_{ensP}(x,x')=max(\gamma,l_P(x,x'))
\end{equation}

\section{Experiments}
In this section, in order to validate the effectiveness of the proposed methods, we generate adversarial examples for fooling classifiers pre-trained on the ImageNet dataset \cite{Deng2009ImageNet} , which consist of 1.2 million natural images collected from Internet and categorized into 1000 classes. We first specify the experimental settings in Sec.4.1. Then we show the results for attacking a single model in Sec.4.2 and an ensemble of model in Sec.4.3.

\subsection{Experiment setting}
We use eleven models, nine of which are normally trained models--Inception V3(Inc-v3)\cite{Szegedy2016Rethinking}, Inception V4(Inc-v4)\cite{Szegedy2016Inception}, Inception Resnet V2(IncRes V2)\cite{Szegedy2016Inception}, Resnet V2-101(Res-101) \cite{He2015Deep} , PolyNet\cite{Zhang2017PolyNet}, SENet154(SENet)\cite{Hu2017Squeeze},PNASNet 5-Large(PNASNet)\cite{Liu2017Progressive}, NASNet-A-Large(NASNet)\cite{Zoph2017Learning}, DenseNet 121(Den-121)\cite{Huang2017Densely}. In order to avoid the evaluate result influenced by resize operation and easy for the ensemble models training, we fine-tune all the models with a $ 299 \times 299$ input by modify the pooling size of last average pooling layer. the other twp models are trained by ensemble adversarial training---Inc-v3$_{ens3}$,Inc-v3$_{ens4}$, as we don't have enough time and resources for prepare these models, we used the models shared by \footnote {https://github.com/dongyp13/Non-Targeted-Adversarial-Attacks} and transfer them from tensorflow model to pytorch model with\footnote{https://github.com/Microsoft/MMdnn}.\par
As we focus on fooling the target white or black box models, we use the fooling rate instead of the attack success rate. We define the fooling rate as how many adversarial examples' prediction labels are different from the origin images' prediction label. Because the classifier is not correct for all the input, in most cases, the attack success rate is higher than the fooling rate. We use the DEV imageset released by CAAD to test all of our methods.\par
In our experiments, we compare our methods to FGSM\cite{2014arXiv1412.6572G} (one-step gradient-based) methods, MI-FGSM (iterative methods) and PGD\cite{2018arXiv180205666U}(iterative methods). Meanwhile we also compare with the method we followed --- ATN(based on Autoencoder), as ATN's loss function is designed for targeted attack, we modify the loss for non-targeted by minimize the true label's prediction. Since optimization-based methods cannot explicitly control the distance between the adversarial examples and the corresponding real images, we don't compare with these methods.\par
\subsection{Attacking a single model}

We show the fooling rates of attacks against the models we consider in Sec.4.1 in Table~\ref{tab:single}. The adversarial examples are generated for Inc-v3 useing FGSM, MI-FGSM and PGD and four of our methods:feature loss based ATN(F-ATN), prediction loss based ATN(P-ATN) and both of the models without the Robust-enhance module. Inc-v4, IncRes-v2, PolyNet, NasNet, Res-101, Inc-v3$_{ens3}$, Inc-v3$_{ens4}$ are black-box models for evaluate transfer-ability of all the methods.The maximum perturbation $\epsilon$ is set to 16 among all experiments, with pixel value in [0,255]. The number of iterations is 10 for MIM-FGSM, and the decay factor $\mu$  is 1.0 as used in MIM. The noise mean factor $\beta$ is 6 in both P-ATN and F-ATN\par
From the table we can observe that our two models could attack the white-box model with a near 100\% fooling rate like MI-FGSM and better than FGSM and PGD. But when it comes to the black-box attack, it can be seen that the performance of FGSM, MI-FGSM and PGD are decrease largely, especially when attacking the adversarial trained models Inc-v3$_{ens3}$ and Inc-v3$_{ens4}$, all of the three attack is powerless. But with Robust-enhance module, both of our F-ATN and P-ATN still keep a high fooling rate. The last row of Table~\ref{tab:single} shows the average black-box fooling rate, P-ATN have the best performance.\par
Meanwhile, We also test the performance of Robust-enhance module, P-ATN(No Robust) and F-ATN(No Robust) show the fooling rate of model without Robust-enhance module during training, results show that with Robust-enhance module, P-ATN's performance have a great progress when attacking black-box models, but F-ATN only have a small improvement. \par 
Although our method improve the success rates greatly for black-box attack even the target is adversarial trained, the performance is not good enough(less than 90\%), we will show that with muti-model ensemble training, our methods will get a better result.\par

%\subsubsection{Loss function}

%\begin{figure*} 
%\centering
%\includegraphics[width=6in,height=4in]{fp.jpg}
%\caption{Visual difference between the loss function. Different loss lead to completely different %modify pattern, for P-ATN, the pattern looks like a checkerboard. But for F-ATN, it looks like smeared %by a brush.}
%\label{fig:loss_diff}
%\end{figure*}

%Comparing with ATN, one of our contributions is use two different loss functions to improve the performance of the ATN. Therefore, we compare our loss functions with ATN in this part.\par Figuer~\ref{fig:loss_diff} show the adversarial examples generated by model trained with different loss function. From Table~\ref{tab:single} we find our loss function based on prediction performs better, if we finetune our model by the robust-enhance module, the P-ATN have a great progress in transfer-ability, but F-ATN just have a slightly improvement. 

\subsubsection{Performance related with Robust-enhance module}
\begin{table}
\begin{center}
\begin{tabular}{|l|c|c|c|}
\hline
Roubst method & Inc-v3 & Inc-v3$_{ens3}$  & Inc-v3$_{ens4}$  \\
\hline\hline
None & 0.98 & 0.58 & 0.47 \\
Random Noise    & 0.97    &\textbf{0.83 } & \textbf{0.80} \\
Pre-trained Filter     &0.97 & 0.43  & 0.56\\
Training Filter& 0.97 & 0.59  & 0.43 \\
\hline
\end{tabular}
\end{center}
\caption{Results.  }
\label{tab:robust}
\end{table}

The Robust-enhance module is the most important part for improving our model's transfer-ability and robustness. Therefore, we study the difference between the Random Noise method, Pretrain Filter method and Maxmin Filter method.\par
We attack Inc-V3 model by P-ATN with three only random noise method, pretrained filer combine with noise and Maxmin method. For the noise method, the noise max mean factor $\beta$ is 6. We show the fooling rate of the generated adversarial examples against Inc-v3, Inc-v3$_{ens3}$, Inc-v3$_{ens4}$ in Table~\ref{tab:robust}. Table~\ref{tab:robust} shows the result of different methods and random noise is the best one.
%The Robust-enhance module is the most import part for improving our model's transfer-ability and become robust. Therefore, we study the difference between the Random Noise method, Pretrain Filter method and Maxmin Filter method. Meanwhile, we study the appropriate value of the noise max mean factor $\beta$ to find out the best hyper parameters.\par
%We attack Inc-V3 model by P-ATN with three only random noise method, pretrained filer combine with noise and Maxmin method. For the noise method, the noise max mean factor $\beta$ ranging from 1 to 9with a granularity 1. We show the fooling rate of the generated adversarial examples against Inc-v3, Inc-v3$_{ens3}$, IncRes V2 and PolyNet in Fig.1 and Table.1. Fig.1 shows the transfer-ability influenced by $\beta$ and we found XXX is a good hyper parameter. Table.1 shows the result of different methods and random noise is the best one.

\subsubsection{Performance on attacking model with image resize}
\begin{table}
\begin{center}
\begin{tabular}{|l|c|c|c|c|}
\hline
Attack & Resize & Inc-v3 & IncRes-v2 & Res-101  \\
\hline\hline
FGSM & N &0.71 & 0.28 & 0.34 \\
FGSM & Y &0.46 & 0.21 & 0.30 \\
\hline
PGD & N &0.99 & 0.12 & 0.18 \\
PGD & Y &0.36 & 0.09 & 0.13 \\
\hline
MI-FGSM & N &0.99 & 0.38 & 0.45 \\
MI-FGSM & Y &\textbf{0.63} & 0.25 & 0.34 \\
\hline
P-ATN & N &0.97 & 0.93 & 0.95 \\
P-ATN & Y &0.54 &\textbf{ 0.34} & \textbf{0.72} \\
\hline
\end{tabular}
\end{center}
\caption{Results.   Ours is better.}
\label{tab:resize}
\end{table}
We then study the adversarial examples' robustness when the black-box model use some image preprocess methods to defence attack. We use the same hyper-parameters for all the attack methods as Sec.4.2, and before we feed the adversarial examples to the target model, we resize it from 299 to 399, then from 399 to 199, finally we resize it back to 299. And the attack performance shows in Table~\ref{tab:resize}.\par
The result shows that after the resize option, all the attack performance for white-box attack(Inc-v3) decrease, especially our methods. As for black-box attack(IncRes-v2 and Res-101) results, the gradients based methods' performance just have a slightly decrease and our method's fooling rate are influenced seriously, but still better than other methods.
\subsubsection{Performance related with size of perturbation}
\begin{figure} 
\centering
\includegraphics[width=3.5in,height=2in]{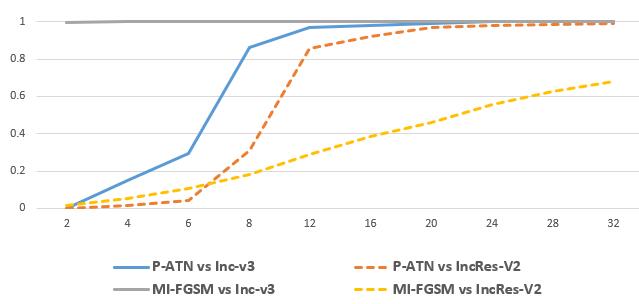}
\caption{}
\label{fig:eps}
\end{figure}
We finally study the influence of the size of adversarial perturbation on the fooling rates. We attack the Inc-v3 model by P-ATN and MI-FGSM with $\epsilon$ from 1 to 32 with a granularity 4 and  the pixels range is [0,255]. We evaluate the attack performance on white-box model Inc-v3, a black-box model IncRes V2. For P-ATN, we use Random Noise module with $\beta=6$ and step size $\alpha$ for PGD and MI-FGSM is 10. As it cost many time for training P-ATN with different epsilon, we just train the model with epsilon 4,8,16,32 and clip to generate different perturbation. \par
Figer.~\ref{fig:eps} show the result. We find that when attacking white-box model Inc-v3, MI-FGSM keeps a high fooling rate for all the epsilon. When the epsilon is small(2,4,6), P-ATN have a poor performance, with the epsilon grow, the fooling rate reach 100\%. When attacking the black-box model, fooling rate of MI-FGSM grow linearly with the size of perturbation, and P-ATN's fooling rate grow exponentially, when the epsilon is large than 8, P-ATN got a better performance and as last reach 99\%. 

%\subsection{Attacking an ensemble models}
\begin{table}
\begin{center}
\begin{tabular}{|l|c|c|c|c|}
\hline
Attack &  IncRes-v2 & Inc-v3$_{ens3}$  \\
\hline\hline
MI-FGSM &0.955 & 0.949 \\
F-ATN   & 0.971 & 0.963 \\
P-ATN   &\textbf{ 0.998} & \textbf{0.997} \\
\hline
\end{tabular}
\end{center}
\caption{Results. }
\label{tab:muti}
\end{table}
%In this section, we will show the experiment results of attacking an ensemble of white-box models. As introduced in Sec.3.4, we ensemble the loss.
%\subsubsection{Comparison of ensmble methods }
%Need More Experiments.
%\subsubsection{Attacking on adversarial trained models}
%Need More Experiments.

\subsection{Competition by ensemble of models}
There are three sub-competitions in Competition on Adversarial Attacks and Defenses 2018 organized by GeekPwn, which are the Non-targeted Adversarial Attack, Targeted Adversarial Attack and Defense Against Adversarial Attack. The organizers provide 1000 ImageNet-compatible images for evaluating the attack and defense submissions. in the non-targeted attack, we won the second place by P-ATN and third place by F-ATN.\par
For both of the network, we used ten pretrained models mentioned in Sec4.1 except Inc-v3. and Robust-enhance module. The noise mean factor $\beta$ is 6 in both P-ATN and F-ATN and we set the ensemble threshold factor $\gamma$ as $-0.9$. For PolyNet and Inc-v3$_{ens3}$, we set the weight as 0.5 and the rest Inc-v3$_{ens4}$, Inc-v4, IncRes V2, Res-101, SENet, PNASNet, NASNet and Den-121, we set the weight as 1.0. Table~\ref{tab:muti} shows the performance comparing with last year's winner's submission .

{\small
\bibliographystyle{ieee}
\bibliography{egbib}

\begin{thebibliography}{10}\itemsep=-1pt

\bibitem{DBLP:journals/corr/BalujaF17}
S.~Baluja and I.~Fischer.
\newblock Adversarial transformation networks: Learning to generate adversarial
  examples.
\newblock {\em CoRR}, abs/1703.09387, 2017.

\bibitem{Deng2009ImageNet}
J.~Deng, W.~Dong, R.~Socher, and L.~J. Li.
\newblock Imagenet: A large-scale hierarchical image database.
\newblock In {\em Computer Vision and Pattern Recognition, 2009. CVPR 2009.
  IEEE Conference on}, pages 248--255, 2009.

\bibitem{2014arXiv1412.6572G}
I.~J. {Goodfellow}, J.~{Shlens}, and C.~{Szegedy}.
\newblock {Explaining and Harnessing Adversarial Examples}.
\newblock {\em ArXiv e-prints}, Dec. 2014.

\bibitem{He2015Deep}
K.~He, X.~Zhang, S.~Ren, and J.~Sun.
\newblock Deep residual learning for image recognition.
\newblock pages 770--778, 2015.

\bibitem{Hu2017Squeeze}
J.~Hu, L.~Shen, and G.~Sun.
\newblock Squeeze-and-excitation networks.
\newblock 2017.

\bibitem{Huang2017Densely}
G.~Huang, Z.~Liu, L.~V.~D. Maaten, and K.~Q. Weinberger.
\newblock Densely connected convolutional networks.
\newblock In {\em IEEE Conference on Computer Vision and Pattern Recognition},
  pages 2261--2269, 2017.

\bibitem{DBLP:journals/corr/LedigTHCATTWS16}
C.~Ledig, L.~Theis, F.~Huszar, J.~Caballero, A.~P. Aitken, A.~Tejani, J.~Totz,
  Z.~Wang, and W.~Shi.
\newblock Photo-realistic single image super-resolution using a generative
  adversarial network.
\newblock {\em CoRR}, abs/1609.04802, 2016.

\bibitem{DBLP:journals/corr/abs-1712-02976}
F.~Liao, M.~Liang, Y.~Dong, T.~Pang, J.~Zhu, and X.~Hu.
\newblock Defense against adversarial attacks using high-level representation
  guided denoiser.
\newblock {\em CoRR}, abs/1712.02976, 2017.

\bibitem{Liu2017Progressive}
C.~Liu, B.~Zoph, M.~Neumann, J.~Shlens, W.~Hua, L.~J. Li, L.~Fei-Fei,
  A.~Yuille, J.~Huang, and K.~Murphy.
\newblock Progressive neural architecture search.
\newblock 2017.

\bibitem{Szegedy2016Inception}
C.~Szegedy, S.~Ioffe, V.~Vanhoucke, and A.~Alemi.
\newblock Inception-v4, inception-resnet and the impact of residual connections
  on learning.
\newblock 2016.

\bibitem{Szegedy2016Rethinking}
C.~Szegedy, V.~Vanhoucke, S.~Ioffe, J.~Shlens, and Z.~Wojna.
\newblock Rethinking the inception architecture for computer vision.
\newblock In {\em Computer Vision and Pattern Recognition}, pages 2818--2826,
  2016.

\bibitem{2018arXiv180205666U}
J.~{Uesato}, B.~{O'Donoghue}, A.~{van den Oord}, and P.~{Kohli}.
\newblock {Adversarial Risk and the Dangers of Evaluating Against Weak
  Attacks}.
\newblock {\em ArXiv e-prints}, Feb. 2018.

\bibitem{Zhang2017PolyNet}
X.~Zhang, Z.~Li, C.~L. Chen, and D.~Lin.
\newblock Polynet: A pursuit of structural diversity in very deep networks.
\newblock In {\em IEEE Conference on Computer Vision and Pattern Recognition},
  pages 3900--3908, 2017.

\bibitem{Zoph2017Learning}
B.~Zoph, V.~Vasudevan, J.~Shlens, and Q.~V. Le.
\newblock Learning transferable architectures for scalable image recognition.
\newblock 2017.

\end{thebibliography}
}

\end{document}